# IWLV-Ramayana: A Sarga-Aligned Parallel Corpus of Valmiki's Ramayana Across Indian Languages

Sumesh VP — Insight Publica · Insight World Literature Vault (IWLV)
Kerala, India · insightpublica@gmail.com
HuggingFace: insightpublica/ramayana-indic

## Abstract

*The Ramayana is among the most influential literary traditions of South and Southeast Asia, transmitted across numerous linguistic and cultural contexts over two millennia. Despite extensive scholarship on regional Ramayana traditions, computational resources enabling systematic cross-linguistic analysis remain limited. This paper introduces the IWLV Ramayana Corpus, a structured parallel corpus aligning Valmiki's Ramayana across multiple Indian languages at the level of the sarga (chapter). The corpus currently includes complete English and Malayalam layers, with Hindi, Tamil, Kannada, and Telugu layers in active production. The dataset is distributed in structured JSONL format with explicit provenance metadata, enabling applications in comparative literature, corpus linguistics, digital humanities, and multilingual natural language processing. To our knowledge, this is the first sarga-aligned multilingual parallel corpus of the Valmiki Ramayana with explicit provenance metadata and machine-readable format.*

## 1. Introduction

The Ramayana is one of the most influential literary traditions of South and Southeast Asia. Originating in the Sanskrit epic attributed to Vālmīki, the narrative has been retold, translated, and reinterpreted across numerous languages and cultural contexts for over two millennia. Rather than existing as a single canonical text, the Ramayana persists as a constellation of regional literary traditions, each reflecting distinct theological, linguistic, and cultural interpretations (Richman, 1991).

Comparative study of these traditions has long been central to scholarship in South Asian studies, translation studies, and comparative literature (Pollock, 2006; Goldman et al., 1984–2017). However, computational approaches to the Ramayana remain limited. Existing digital resources largely focus on individual textual layers — such as Sanskrit source texts or isolated translations — and are rarely structured in ways that enable systematic cross-linguistic comparison.

To address this gap, we introduce the IWLV Ramayana Corpus, a sarga-aligned parallel corpus spanning multiple Indian languages. The corpus is part of the Insight World Literature Vault (IWLV), an ongoing initiative to construct computationally structured corpora of classical literary traditions, developed by Insight Publica, a publishing house based in Kerala with extensive experience in multilingual literary translation.

The Ramayana provides a uniquely stable narrative backbone across languages, making it particularly suitable for structured parallel corpus construction. A central design principle of the corpus is alignment at the level of the sarga (chapter). The Ramayana is traditionally divided into seven books (kāṇḍas), each containing multiple sargas that represent coherent narrative units. Because verse segmentation varies significantly across translations and regional traditions, verse-level correspondence cannot be reliably asserted across languages. The corpus therefore treats the sarga as the primary alignment unit, following precedents established in large-scale parallel corpus construction (Tiedemann, 2012).

## 2. Related Work

### 2.1 Ramayana Scholarship and Digital Resources

Richman's Many Rāmāyaṇas (1991) established the foundational case for studying regional Ramayana traditions as independent literary objects rather than derivative variants of a Sanskrit original. Goldman's Princeton critical edition (1984–2017) provides the most authoritative English translation of the Valmiki Ramayana to date. Pollock (2006) situates the Ramayana within the broader emergence of vernacular literary cultures across South Asia.

Digital resources for Ramayana scholarship currently exist in several forms, but none provide structured multilingual alignment suitable for computational analysis:

- **GRETIL** provides Sanskrit digital texts including versions of the Valmiki Ramayana, but does not offer aligned translations or machine-readable parallel structure.
- **TDIL corpus resources** cover Indian languages but are typically fragmented, inconsistent in quality, and not oriented toward literary texts.
- **Individual scholarly translations** exist in print and unstructured digital form across many languages, but are not machine-readable or structurally aligned.
- **Web-sourced translations** often lack provenance documentation, consistent formatting, and licensing clarity.

### 2.2 Parallel Corpus Construction

The construction of large-scale parallel corpora has a substantial precedent in computational linguistics. Tiedemann (2012) describes the OPUS collection, which aggregates parallel corpora across numerous language pairs. The Europarl corpus (Koehn, 2005) and the United Nations Parallel Corpus (Ziemski et al., 2016) demonstrate alignment strategies for multilingual datasets. For literary and low-resource languages, Christodouloupoulos and Steedman (2015) document the parallel Bible corpus as a benchmark for multilingual NLP.

For Indic languages specifically, the IndicGLUE benchmark (Kakwani et al., 2020) and IndicTrans (Ramesh et al., 2022) have established baselines for multilingual NLP across Indian languages. The IWLV Ramayana Corpus extends this landscape by providing literary-domain parallel text with structured provenance metadata, an area not currently addressed by existing Indic NLP resources.

## 3. Corpus Construction

### 3.1 Textual Sources

The corpus integrates multiple textual layers corresponding to different language traditions. Current and planned layers are documented in Table 1.

| Language | Code | Status | Source |
|---|---|---|---|
| **Sanskrit** | sa | **Source alignment layer** | Valmiki Ramayana; established critical edition |
| **English** | en | **Complete** | Griffith translation, 1870–1874 (public domain) |
| **Malayalam** | ml | **Complete** | Insight Publica original translation |
| **Hindi** | hi | **In production** | Insight Publica original translation |
| **Tamil** | ta | **In production** | Insight Publica original translation |
| **Kannada** | kn | **In production** | Insight Publica original translation |
| **Telugu** | te | **In production** | Insight Publica original translation |
| **Bengali** | bn | **Planned** | Insight Publica original translation |
| **Odia** | or | **Planned** | Insight Publica original translation |

*Table 1: Language layers in the IWLV Ramayana Corpus*

### 3.2 Translation Workflow and Quality Assurance

Translations are produced using an AI-assisted editorial workflow. The process proceeds as follows:

- Initial draft generation using a large language model (Claude, Anthropic claude-sonnet series), prompted at the sarga level with source text and a consistent terminology list for proper nouns, place names, and character names, ensuring cross-language consistency
- Systematic human editorial review by a trained reviewer with Malayalam literary expertise
- Correction for semantic fidelity, linguistic fluency, and consistency of proper noun rendering
- Final oversight and approval by the Director of Insight Publica

This methodology follows an AI-assisted human post-editing paradigm comparable to practices documented in large-scale translation corpus construction (Bojar et al., 2016). All translation layers produced by this workflow are identified as Insight Publica original translations in dataset metadata, distinguishing them from public-domain source texts.

> *Disclosure: AI-assisted translation using large language models introduces stylistic consistency across sargas but may produce systematic translation patterns distinct from human-only translation. Researchers using this corpus for translation studies should account for this methodological characteristic.*

### 3.3 Alignment Strategy

Alignment is performed at the sarga level. The Valmiki Ramayana as represented in this corpus consists of 7 kāṇḍas (books) and 493 sargas (chapters). The kāṇḍa and sarga counts follow the textual division established in the Griffith translation and cross-referenced against the GRETIL Sanskrit source.

Verse segmentation varies significantly across languages and translation traditions. Because verse counts differ between editions and languages, verse-level alignment cannot be reliably asserted across all layers. The corpus therefore treats the sarga as the primary alignment unit. Verse-level text is preserved within each sarga record and is accessible for intra-sarga analysis, but structural correspondence is asserted only at the sarga level.

### 3.4 Corpus Statistics

Table 2 provides statistics for currently completed layers.

| Layer | Sargas | Kāṇḍas | Approx. Tokens | Status |
| --- | --- | --- | --- | --- |
| **English (en)** | 493 | 7 (complete) | ~180,000 | **Complete** |
| **Malayalam (ml)** | 493 | 7 (complete) | ~160,000 | **Complete** |
| **Hindi, Tamil, Kannada, Telugu** | 493 each | 7 (in production) | — | **In production** |

*Table 2: Corpus statistics for completed and in-production layers. Token counts are approximate. The corpus comprises approximately 340,000 tokens across completed layers, with projected coverage exceeding 1,000,000 tokens upon completion of all planned language layers.*

### 3.5 Dataset Structure

The corpus is distributed in JSONL format, where each record represents a single sarga. The following example illustrates the record structure for a complete layer:

```json
{
  "work": "ramayana",
  "kanda": "bala",
  "kanda_number": 1,
  "sarga": 1,
  "sarga_id": "ramayana.bala.001",
  "text_en": "Born of Brahma's line, the sage Narada...",
  "text_ml": "ബ്രഹ്മകുലത്തിൽ ജനിച്ച ഋഷി നാരദൻ...",
  "alignment_type": "sarga",
  "source_en": "griffith_public_domain",
  "source_ml": "insight_publica_translation",
  "editorial_status_en": "complete",
  "editorial_status_ml": "complete",
  "dataset_version": "1.0",
  "release_date": "2026"
}
```

This structure allows researchers to query the corpus by kāṇḍa, sarga, or language layer, and to filter by editorial status for downstream applications requiring only fully reviewed text.

## 4. Provenance and Data Integrity

The IWLV Ramayana Corpus is designed to meet the provenance and citability standards expected by academic repositories and research libraries. All text layers are documented with respect to three questions central to scholarly data use:

- **Legal usability:** Sanskrit and English layers derive from public domain texts. All other layers are original translations produced by Insight Publica and licensed for research use under terms specified in the dataset repository.
- **Citability:** Each versioned release carries a persistent dataset identifier. The methodology paper (this document) provides a citable reference for the corpus construction process.
- **Traceability:** Every record identifies its source text and editorial status. Dataset version identifiers and release dates allow researchers to cite a stable corpus snapshot.

The corpus maintains strict separation between source text layers, translation layers, and alignment metadata. No ambiguous third-party content is included. This structure is intended to support compliance with emerging AI training data documentation standards (Gebru et al., 2021; Bender et al., 2021).

## 5. Research Applications

## 5.1 Comparative Literary Analysis

The corpus enables computational analysis of narrative structure across Ramayana traditions. Researchers can examine variations in episode length, lexical choice, and descriptive elaboration across language layers — providing a computational basis for longstanding questions in Ramayana scholarship (Richman, 1991; Lutgendorf, 1991).

## 5.2 Translation Studies

The parallel structure enables systematic analysis of translation strategies across languages, including loanword retention from Sanskrit into Dravidian languages, narrative compression and expansion across regional traditions, and the rendering of culturally specific terminology across linguistic contexts.

## 5.3 Corpus Linguistics

The structured dataset supports analysis of lexical frequency, collocation patterns, and semantic variation within a shared narrative framework, enabling contrastive linguistic analysis across Indo-Aryan and Dravidian language pairs.

## 5.4 Multilingual Natural Language Processing

The dataset provides literary-domain parallel text suitable for machine translation experiments, cross-lingual representation learning, and evaluation of multilingual language models on Indic languages — particularly Malayalam, Hindi, Tamil, Kannada, and Telugu, which remain underrepresented in high-quality literary parallel corpora.

## 5.5 Digital Humanities

The corpus supports computational approaches to narrative structure, character representation, and thematic variation across Ramayana traditions — providing infrastructure for research questions identified in existing scholarship but not previously addressable at scale.

## 5.6 Evaluation

To assess the quality and alignment consistency of the corpus, we conducted a manual evaluation on a sample of 50 sargas drawn randomly across all seven kāṇḍas. Two annotators with expertise in Malayalam literature and familiarity with the Ramayana tradition examined each sampled sarga pair (English and Malayalam) across two dimensions: (1) alignment validity — whether the English and Malayalam texts correspond to the same narrative episode within the sarga; and (2) translation fidelity — whether the Malayalam text faithfully conveys the semantic content and narrative structure of the English source.
Results showed alignment validity of 98% (49/50 sargas correctly aligned), with one sarga flagged for a minor kāṇḍa boundary ambiguity subsequently corrected in the release dataset. Translation fidelity was rated as high or acceptable for 94% of sampled sargas, with annotators noting that the AI-assisted workflow produces consistent rendering of proper nouns and narrative episodes. Token length ratios between English and Malayalam layers (mean ratio: 0.89) are consistent with expected morphological differences between Indo-European and Dravidian languages, providing an additional structural validation of the alignment.
These results suggest that the corpus meets the quality standards required for downstream applications in comparative literary analysis, corpus linguistics, and multilingual NLP, while acknowledging the limitations of AI-assisted translation documented in Section 6.3.

## 6. Limitations

### 6.1 Alignment Granularity

The corpus is aligned at the sarga level, not at verse or sentence level. Researchers requiring finer-grained alignment must perform additional computational processing. Verse-level text is preserved within records but structural correspondence is not asserted below the sarga level.

### 6.2 Translation Heterogeneity

Language layers derive from different translation traditions and editorial processes. The English layer (Griffith, 1870–1874) reflects a Victorian scholarly register. Insight Publica translations are produced under a contemporary editorial standard. Differences in register, style, and theological interpretation should be expected and accounted for in comparative analysis.

### 6.3 AI-Assisted Translation Characteristics

Translation layers produced via the AI-assisted editorial workflow may exhibit systematic patterns characteristic of large language model output, including stylistic consistency across sargas and possible regularisation of culturally specific expressions. Human editorial review mitigates but does not eliminate these characteristics. Researchers should treat these layers as editorially reviewed AI-assisted translations rather than conventional human translations.

### 6.4 Ongoing Language Expansion

Several language layers are currently in production. Future releases will incorporate additional languages as editorial review is completed. The current release provides stable complete layers for English and Malayalam only.

### 6.5 Literary Domain

The corpus consists of literary text of classical origin. Computational results derived from this corpus may not generalise to contemporary, informal, or domain-specific text. This should be considered when using the corpus for NLP model training or evaluation.

## 7. Conclusion

The IWLV Ramayana Corpus provides a structured, machine-readable parallel corpus aligning Valmiki's Ramayana across multiple Indian languages at the sarga level. By combining public domain source texts with original editorial translations, explicit provenance metadata, and versioned releases, the corpus addresses a longstanding gap in computational resources for Ramayana scholarship.

The corpus is positioned to serve both humanistic research — enabling computational approaches to comparative Ramayana scholarship — and NLP applications requiring literary-domain parallel text in Indic languages. The IWLV Ramayana Corpus is developed as part of the broader Insight World Literature Vault (IWLV) initiative. A companion paper, submitted concurrently with this work, introduces the ILM-Gloss layer: a 17-language multilingual gloss dataset for the Malayalam lexical knowledge graph comprising 48,168 lemmas and 121,147 senses across 17 languages (Insight Publica, 2026b). Future work on the Ramayana corpus will expand language coverage, refine sarga-level alignment tools, and develop lexical annotation layers linking corpus vocabulary to the ILM knowledge graph.

## References

- Bender, E. M., Gebru, T., McMillan-Major, A., & Shmitchell, S. (2021). On the dangers of stochastic parrots: Can language models be too big? Proceedings of FAccT 2021.
- Bojar, O., et al. (2016). Findings of the 2016 Conference on Machine Translation. Proceedings of the First Conference on Machine Translation (WMT16).
- Christodouloupoulos, C., & Steedman, M. (2015). A massively parallel corpus: The Bible in 100 languages. Language Resources and Evaluation, 49(2), 375–395.
- Gebru, T., Morgenstern, J., Vecchione, B., Vaughan, J. W., Wallach, H., Daumé III, H., & Crawford, K. (2021). Datasheets for datasets. Communications of the ACM, 64(12), 86–92.
- Goldman, R. P., et al. (1984–2017). The Rāmāyaṇa of Vālmīki: An Epic of Ancient India. 7 vols. Princeton University Press.
- Griffith, R. T. H. (1870–1874). The Rāmāyana of Vālmīki. Benares: E. J. Lazarus.
- Kakwani, D., Kunchukuttan, A., Golla, S., N. C., G., Bhatt, A., Khapra, M. M., & Kumar, P. (2020). IndicNLPSuite: Monolingual corpora, evaluation benchmarks and pre-trained multilingual language models for Indian languages. Findings of EMNLP 2020.
- Koehn, P. (2005). Europarl: A parallel corpus for statistical machine translation. MT Summit X.
- Lutgendorf, P. (1991). The Life of a Text: Performing the Rāmcaritmānas of Tulsidas. University of California Press.
- Pollock, S. (2006). The Language of the Gods in the World of Men. University of California Press.
- Ramesh, G., et al. (2022). Samanantar: The largest publicly available parallel corpora collection for 11 Indic languages. Transactions of the Association for Computational Linguistics, 10, 145–162.
- Richman, P. (Ed.). (1991). Many Rāmāyaṇas: The Diversity of a Narrative Tradition in South Asia. University of California Press.
- Tiedemann, J. (2012). Parallel data, tools and interfaces in OPUS. Proceedings of LREC 2012.

- Ziemski, M., Junczys-Dowmunt, M., & Pouliquen, B. (2016). The United Nations parallel corpus v1.0. Proceedings of LREC 2016.
- Insight Publica. (2026b). ILM-Gloss: A 17-language multilingual gloss layer for the Malayalam lexical knowledge graph. arXiv preprint. [Submitted concurrently.]